\pdfoutput=1

\documentclass[11pt]{article}

\usepackage[]{acl}

\usepackage{times}
\usepackage{latexsym}

\usepackage[T1]{fontenc}

\usepackage[utf8]{inputenc}

\usepackage{microtype}

\usepackage{amsmath}
\usepackage{xspace}
\usepackage{listings}
\usepackage{xcolor}
\usepackage{booktabs}
\usepackage{lipsum}
\usepackage{array}
\usepackage{multirow}
\usepackage{enumitem}
\usepackage{amssymb}
\usepackage{pifont}
\newcommand{\cmark}{\ding{51}}
\newcommand{\xmark}{\ding{55}}

\usepackage{booktabs}

\definecolor{codegreen}{rgb}{0,0.6,0}
\definecolor{codegray}{rgb}{0.5,0.5,0.5}
\definecolor{codepurple}{rgb}{0.58,0,0.82}
\definecolor{backcolour}{rgb}{0.95,0.95,0.92}

\lstdefinestyle{mystyle}{
    backgroundcolor=\color{backcolour},   
    commentstyle=\color{codegreen},
    keywordstyle=\color{magenta},
    numberstyle=\tiny\color{codegray},
    stringstyle=\color{codepurple},
    basicstyle=\ttfamily\footnotesize,
    breakatwhitespace=false,         
    breaklines=true,                 
    captionpos=b,                    
    keepspaces=true,                 
    numbers=left,                    
    numbersep=5pt,                  
    showspaces=false,                
    showstringspaces=false,
    showtabs=false,                  
    tabsize=2,
    escapeinside={(*@}{@*)}
}

\lstdefinestyle{tabstyle}{
    backgroundcolor=\color{backcolour},   
    commentstyle=\color{codegreen},
    keywordstyle=\color{magenta},
    numberstyle=\tiny\color{codegray},
    stringstyle=\color{codepurple},
    basicstyle=\ttfamily\footnotesize,
    breakatwhitespace=false,         
    breaklines=true,                 
    captionpos=b,                    
    keepspaces=false,                 
    numbers=none,                    
    numbersep=5pt,                  
    showspaces=false,                
    showstringspaces=false,
    showtabs=false,                  
    tabsize=2,
    escapeinside={(*@}{@*)}
}

\lstdefinestyle{tabstylefordialogue}{
    backgroundcolor=\color{backcolour},   
    commentstyle=\color{codegreen},
    keywordstyle=\color{magenta},
    numberstyle=\tiny\color{codegray},
    stringstyle=\color{codepurple},
    basicstyle=\ttfamily\scriptsize,
    breakatwhitespace=false,         
    breaklines=true,                 
    captionpos=b,                    
    keepspaces=false,                 
    numbers=none,                    
    numbersep=5pt,                  
    showspaces=false,                
    showstringspaces=false,
    showtabs=false,                  
    tabsize=2,
    escapeinside={(*@}{@*)}
}

\usepackage{graphicx}

\newcommand{\EMR}{EMR\xspace}

\newcommand{\nikos}[1]{}{}
\newcommand{\codex}[1]{#1}

\usepackage{xargs}
\usepackage[colorinlistoftodos,prependcaption,textsize=tiny]{todonotes}
\newcommandx{\comment}[2][1=]{\todo[linecolor=red,backgroundcolor=red!25,bordercolor=red,#1]{#2}}
\newcommandx{\xcomment}[2][1=]{\todo[linecolor=blue,backgroundcolor=blue!25,bordercolor=blue,#1]{#2}}
\newcommandx{\xxcomment}[2][1=]{\todo[disable,#1]{#2}}

\title{Dialog2API: Task-Oriented Dialogue with API Description and Example Programs}

\author{
    Raphael Shu\quad
    Elman Mansimov\quad 
    Tamer Alkhouli\quad
    Nikolaos Pappas\quad
    Salvatore Romeo\quad\\
    \bf{
    Arshit Gupta\quad
    Saab Mansour\quad
    Yi Zhang\quad
    Dan Roth
    } \\
    AWS AI Labs\\
    \texttt{\small{\{zhongzhu,mansimov,alkhouli,nppappa,romeosr,arshig,saabm,yizhngn,drot\}@amazon.com}}
}

\begin{document}
\maketitle
\begin{abstract}
\xxcomment{DR: I think that the abstract does not explain well the challenges addressed and the innovative solution. It's good to start with the limitation of the current state; maybe even a concise example; but then, it is not clear why and how Dialog2API solves it; maybe the initial description needs to be at a higher level of abstraction to introduce the notion of generation programs that interact with APIs?}
\xxcomment{YZ: I agree with Dan's comments. With Dialog2API, we are introducing a new paradigm to task-oriented dialogue. Dialog2API strikes a middle ground between fixed/closed-schema TOD (intent/slot-based, or semantic parsing based) and fully open-schema TOD (through full-on program synthesis such as SMCalFlow). Dialog2API generates executable wrapper code around a closed set of task APIs, allowing for flexible and robust goal-oriented multi-step interaction between user and system.}

Functionality and dialogue experience are two important factors of task-oriented dialogue systems. Conventional approaches with closed schema (e.g., conversational semantic parsing) often fail as both the functionality and dialogue experience are strongly constrained by the underlying schema.

\xxcomment{DR: Why do we call it task rather than a paradigm for developing dialogue systems?}
We introduce a new paradigm for task-oriented dialogue -- Dialog2API -- to greatly expand the functionality and provide seamless dialogue experience. The conversational model interacts with the environment by generating and executing programs triggering a set of pre-defined APIs. The model also manages the dialogue policy and interact with the user through generating appropriate natural language responses.  By allowing generating free-form programs, Dialog2API supports composite goals by combining different APIs, whereas unrestricted program revision provides natural and robust dialogue experience.

\xxcomment{DR: Rather than "task input" I think that we are saying that "in order to facilitate Dialog2API, the model (should we call it "dialog manager, or is this taken already?) needs to be able to generate programs...; we make use of large LLMs fine tuned with a small set of API documents to induce such programs...}

To facilitate Dialog2API, the core model is provided with API documents, an execution environment and optionally some example dialogues annotated with programs.
We propose an approach tailored for the Dialog2API, where the dialogue states are represented by a stack of programs, with most recently mentioned program on the top of the stack.

\xxcomment{DR: Can we go beyond just S3? this might not be enough for a strong paper}

Dialog2API can work with many application scenarios such as software automation and customer service. In this paper, we construct a dataset for AWS S3 APIs and present evaluation results of in-context learning baselines.

\end{abstract}

\section{Introduction}

\begin{figure}[!t]
    \centering
    \includegraphics[scale=0.55]{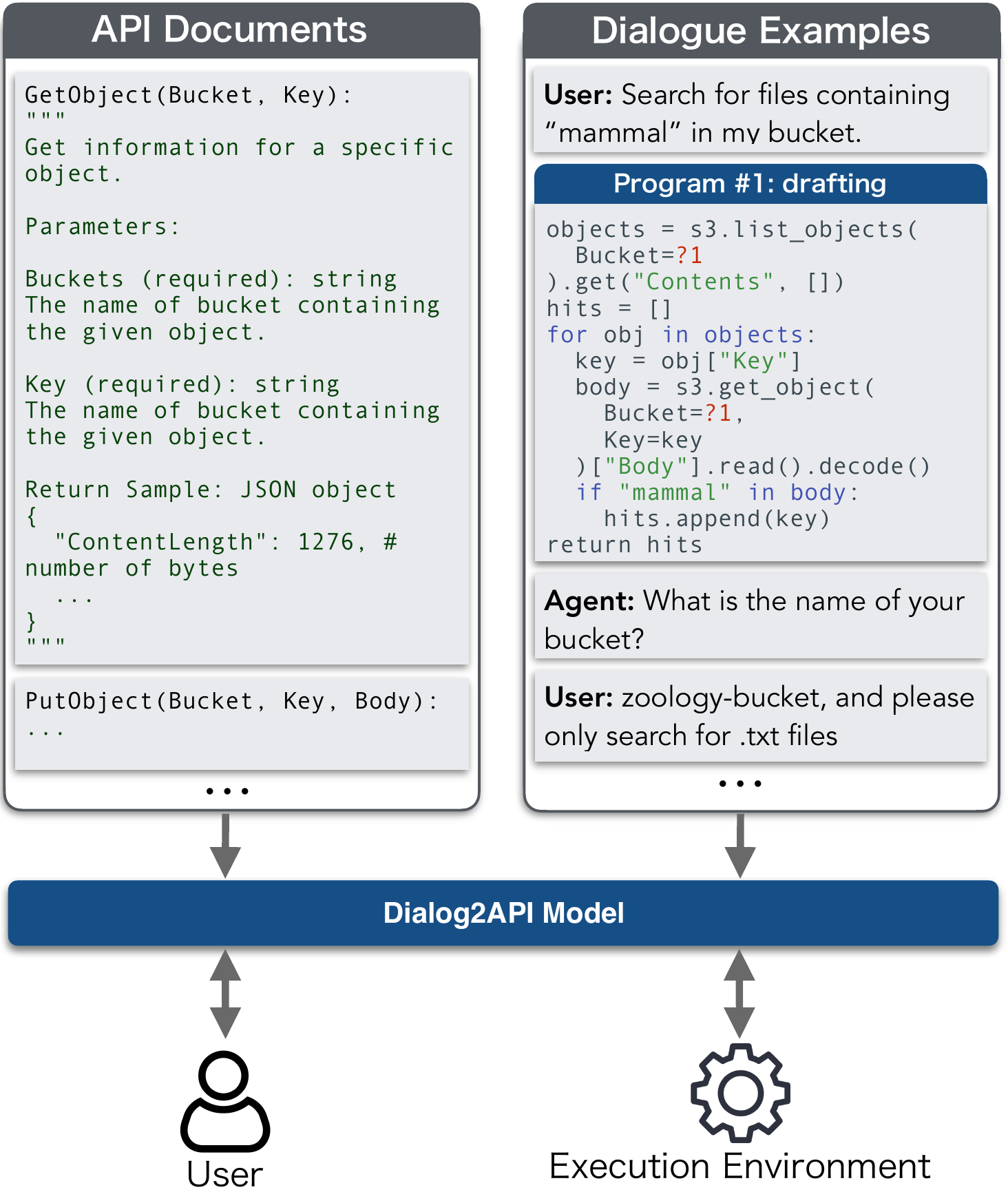}
    \caption{Illustration of Dialog2API paradigm. A chatbot is defined by (1) a set of API definitions and (2) example dialogs with program annotation. The model learns to achieve certain goals by interacting with the user and the execution environment.}
    \label{fig:task}
\end{figure}

Task-oriented dialogue (TOD) systems have obtained wide adoption in industrial applications
\xxcomment{DR: On the results: since the focus is on the conversation and not the program generation, we must highlight that. Ideall, we should present results that (1) illustrate the power of the dialog over 1-turn, and (2) illustrate the power of the new paradigm over the old one; how can we do that?}
\xxcomment{YZ: On 1-turn vs. multi-turn comparison, there's a plan to conduct a human-study to show the benefits of conversation over 1-turn. On the comparison between Dialog2API paradigm vs. traditional approach, we can potentially build a Lex-style or NL2CLI intent/slot based bot and do a comparison study.}
with the help of continuously improved models \cite{wen-etal-2017-network, yang2020, hosseiniasl2020simple, su-etal-2022-multi, valizadeh-parde-2022-ai, zhang2020}. The conventional TOD systems are constructed with a pipeline containing different modules such as natural language understanding, dialogue state tracking, dialogue policy management and response generation \cite{smith1995,Young2013POMDPBasedSS}.

\begin{table*}[!t]
\scriptsize
\centering
\begin{tabular}{ r | c | c | c} \toprule
 & {\small Slot-filling} & {\small Semantic Parsing} & {\small Dialog2API} \\
\midrule
Grounding based on API documents & \xmark & \xmark & \cmark \\
Customizable dialogue experience through example dialogues & \xmark & \xmark & \cmark \\
Flexible goal revision and error recovery & \xmark & \xmark & \cmark\\
Functionality is not constrained by the schema & \xmark & \xmark & \cmark \\
Does not require large amount of annotated data for supervised learning & \cmark & \xmark & \cmark \\
High interpretablility on dialogue policy & \cmark & \cmark & \xmark \\
High controllability on bot response & \cmark & \cmark & \xmark\\
\bottomrule
\end{tabular}
\caption{Comparing the pros and cons of different dialogue paradigms. Dialog2API has a strength on functionality and natural dialogue experience. However, as being more model-dependent, Dialog2API has lower interpretability on dialogue policy and controllability on bot response.}
\label{table:paradigm_comparison}
\end{table*}

To build a task-oriented dialogue system, many developers have followed the slot-filling approach \cite{Pieraccini1992ASU}, where a chatbot is defined using intents and slots are described by sample utterances. Such an approach allow developers to easily run the chatbots with accurate intent and slot classifiers obtained through supervised learning. However, the low extensibility of the functionality in slot-filling approach is becoming a bottleneck for complex and dynamic tasks. For example, a simple food ordering system typically fails when the user says ``Please replace the pizza in my order with pasta and roasted carrot'', as the system have ``order'' and ``cancel'' intents but not a ``replace'' intent. 

Semantic parsing approaches \cite{Gupta2018SemanticPF,Rongali2020DontPG, Aghajanyan2020ConversationalSP} build a hierarchical representation, allowing the chatbot to understand complex and compositional user goals. However, such an approach typically require a large amount of training samples. User request in a form that is not covered by the training data is likely to result in degraded parses. 


To further expand the adoption of TOD systems for complex use cases, we propose the Dialog2API paradigm. The main motivation is to greatly expand the functionality and improve dialogue experience while keeping it easy for building such a system. As illustrated in Fig.~\ref{fig:task}, the developer defines a chatbot by providing two types of information: (1) the API document, and, optionally,  (2) example dialogues with program annotations to customize the dialogue experience. A dialogue model is then required to accomplish certain domain-specific goals by interacting with the user, and the execution environment though API calls. More specifically, the model is required to perform program synthesis, dialogue policy management and response generation. As APIs and corresponding documents are readily available for many applications, the effort required from developers is minimal.

In the proposed paradigm, we ask the model to generate executable programming codes such as Python for achieving the user goals. Given the flexibility on program synthesis, we propose an execution-based metric to evaluate the model performance by matching the environment state rather than the program itself. By measuring the ratio of successful turns in a dialogue, it reflects the quality of real user experience. A summary of pros and cons comparing Dialog2API with existing paradigms can be found in Table~\ref{table:paradigm_comparison}. We show example dialogues with different paradigms in appendix \ref{sec:appendix_paradigm_examples}. The main drawbacks of Dialog2API are on lower interpretability on dialogue policy and controllability on bot response. As Dialog2API does not require maintaining an explicit and discrete dialogue policy, an additional model might be required for explaining the policy. For bot response, as it is generated by the core model, an extra post-processing mechanism should be in place for validating the correctness and guaranteeing the safety. We leave these two items for future research.

Note that the performance of Dialog2API is not solely depending on strong code generation capability. The code has to be fully grounded on the underlying APIs. Moreover, Dialog2API requires the core model to handle conversational complexities for user interaction.


Tailored for the Dialog2API, we propose a novel approach to perform the task-oriented dialogue. We represent the dialogue state with a stack of programs. Each program in the stack corresponds to a sub-goal of the conversation. Most recently mentioned program is placed on the top of the stack.
Each program can have three states: (1) \texttt{drafting}, (2) \texttt{final} and (3) \texttt{abandoned}. A program is immediately executed when the status turns from drafting to final.
The execution outcomes including errors will be attached to the corresponding program in the stack. In this framework, a dialogue model needs to synthesize the updated programs based on the current program stack and dialogue history. Once the program is updated with execution result, the model is required to generate the response.

In this paper, we construct and plan to release a dataset containing 100 conversations for manipulating AWS S3 APIs with turn-level program annotation, and intend to release it. Benchmarking results with in-context learning baselines show that the best baseline reaches 33\% on the proposed execution-based metric. The results also indicate that API documents are helpful on solving the tasks even when the model has access to example dialogues.

\section{Dialog2API}

The Dialog2API paradigm provides a framework to allow the core model to progressively translate the underlying goals of a user to executable programs through dialogue. In each user turn, the Dialog2API model interacts with the user and environment based on three types of the inputs: (1) the API document, (2) the history of current dialogue, and, optionally, (3) example dialogues with program annotations.

The API document should cover all APIs for interacting with the execution environment, with a format designed by the developer. The document provides information for the Dialog2API model to understand the functionality and usage of different APIs, thus helping the generation of valid programs.

The example dialogues demonstrate how the APIs are triggered in the dialogue context and how the chatbot should interact with the user. For each user turn, we require the developer to annotate each underlying user goal with a program \footnote{It is possible for a user turn be annotated with multiple programs, each program for a sub-goal. The programs will be executed sequentially.}. To be developer-friendly, we do not restrict the structure or style of programs as in \citet{Andreas2020TaskOrientedDA}. We allow the programs to be annotated with natural programming codes (e.g., standard Python) as long as the code can be executed. As our evaluation metric is not based on exact program matching, we also do not require programs of different dialogues to share the same programming style, naming convention, and so on.

Each program needs to be assigned a goal ID and the status. There are three possible statuses:
\begin{enumerate}[noitemsep]
    \item \textbf{\texttt{drafting:}} The program is partially or fully generated, but not ready to execute. A program in drafting stage can be revised in a later turn.
    \item \textbf{\texttt{final:}} The program is finalized. Once a program becomes final, it will be immediately executed by the execution environment. Execution results (i.e., return values and errors) will be observed by the Dialog2API model. A final program cannot be further revised.
    \item \textbf{\texttt{abandoned:}} The underlying goal of the program is abandoned by the user.
\end{enumerate}

We denote the API definitions by $\mathcal{D}$, the user utterance, programs, execution results and bot response for turn $t$ by $U_t, P_t, O_t, R_t$ respectively, whereas the dialogue history before a given turn by $H_{<t} = \{U_{<t}, P_{<t}, O_{<t}, R_{<t}\}$. The Dialog2API task requires a model to perform two tasks for each user turn: (1) program prediction: $f(\mathcal{D}, H_{<t}, U_t) \rightarrow P_t$ and (2) response generation: $f(\mathcal{D}, H_{<t}, U_t, P_t, O_t) \rightarrow R_t$. For each program in $P_t$, the model is required to generate its goal ID, status and programming code. The bot response $R_t$ is a natural language utterance.

\subsection{Evaluation Metrics}
\label{sec:eval}

Evaluating the correctness of generated programs in Dialog2API is non-trivial. As we only require the programs to be valid programming codes, there might exist multiple solutions for achieving the same goal. For example, to rename files, instead of using the ``move`` API, the goal is also achievable with ``copy'' and ``delete'' APIs.

In this work, we propose to measure the program prediction performance through \textit{execution match ratio (\EMR)}. To compute \EMR, we require the environment state to be deterministic after executing a program, and a signature (i.e. checksum) of the environment state can be produced. Depending on the domain, an environment can be a database, file system, etc. Intuitively, \EMR is an execution-based metric that measures the ratio of successful turns in a dialogue. The user can be also considered as part of the environment since the return value of a program impacts the dialogue flow by influencing the agent response and thus user observation. Therefore, we also consider the program return values when computing the environment signature.

During evaluation, we generate the programs and agent responses in an end-to-end fashion. In each turn, we feed the model with the corresponding ground-truth user utterance regardless of the previous generated programs and responses. Therefore, the number of turns in the generated dialogue always matches the ground-truth dialogue. We consider a dialogue turn as successful if (1) the Dialog2API model executes a program at the given turn, (2) the environment signature matches the ground-truth signature. More concretely, \EMR is computed as the following ratio:
\begin{gather}
    \mathrm{EMR} = {t_{\mathrm{last}}} / {T}, \\
    t_{\mathrm{last}} = \max \{ t': E(\hat P_{t'}) = E(P_{t'}) \quad \forall t' \in [1, t] \}, \nonumber
\end{gather}
where $T$ is the total number of turns, $P_{t'}$ and $\hat P_{t'}$ are the ground-truth and executed programs respectively in turn $t'$, and $t_{\mathrm{last}}$ is the very last turn that we know the executed program applies same impact to the environment as the ground-truth. $E(\hat P_{t'})$ denotes the environment signature after executing the predicted program in turn $t'$.

In addition to \EMR for evaluating program prediction performance, we also measure the response quality by reporting BLEU scores \cite{papineni-etal-2002-bleu} of generated agent responses based on the ground-truth in the end-to-end setting. As a supplementary metric, we also report the code edit distance\footnote{The programming codes are tokenized using Code Tokenizer (\url{https://pypi.org/project/code-tokenize/}).} between the generated and ground-truth codes in teacher-forced setting (i.e., all inputs to the model are from ground-truth).

\begin{figure}[t]
    \centering
    \includegraphics[scale=0.5]{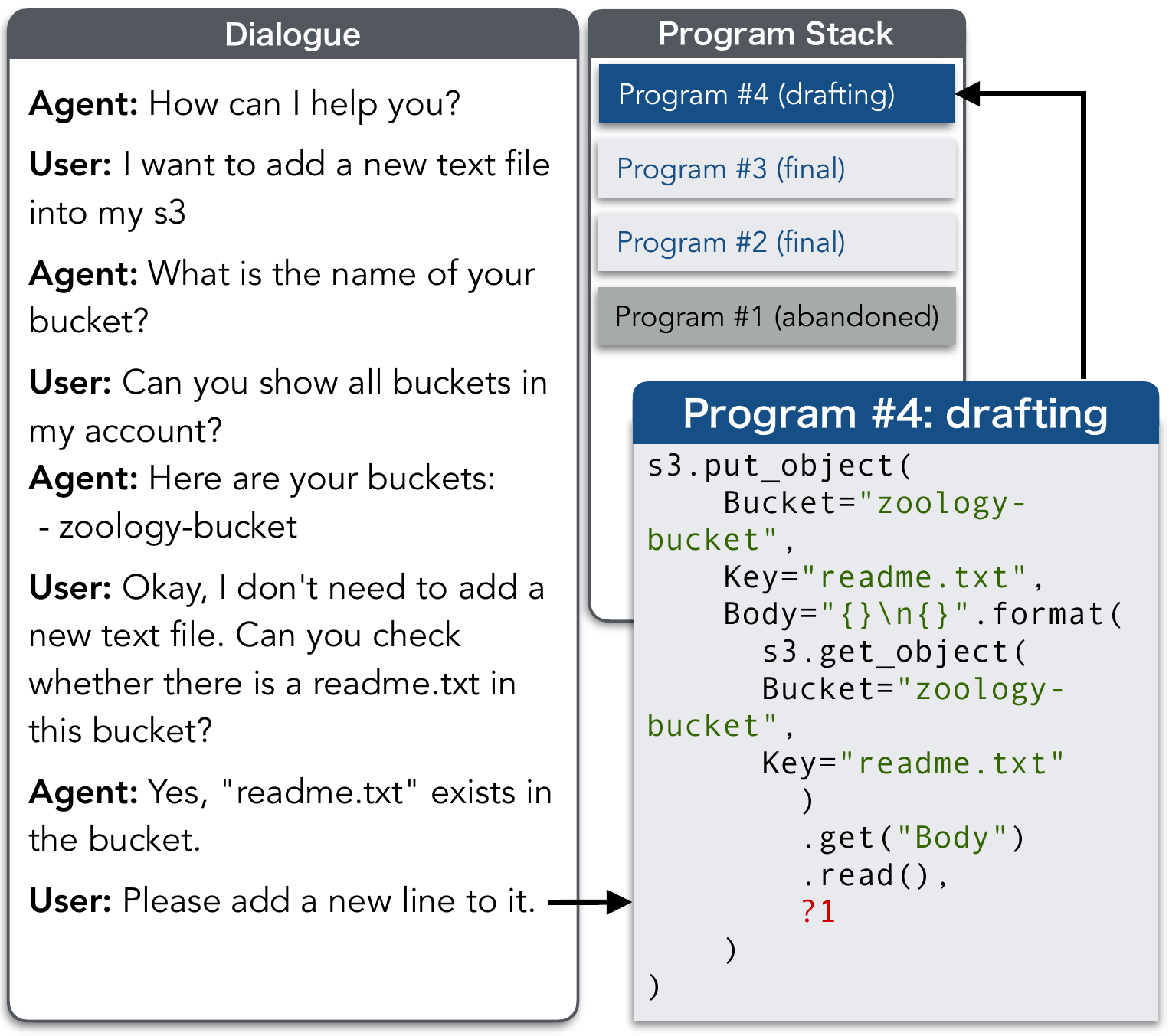}
    \caption{An example program stack after a conversation consisting of eight turns.}
    \label{fig:approach}
\end{figure}

\paragraph{Limitation} The design of \EMR is for approximating the dialogue-level success rate by estimating the percentage of dialogue turns with successful user-bot interaction. This metric can over-estimate or under-estimate in certain cases. As the Dialog2API model is fed with the ground-truth user utterances regardless of the generated agent response, if the model achieves the goal with a different dialogue flow (e.g., when copying a file, confirm the target location first vs. confirm the source location first), the \EMR score can under-estimate. However, also due to such nature, when the model generates correct programs but wrong responses, if the mistake in response does not negatively impact the program prediction in later turns, \EMR can over-estimate\footnote{An over-estimating example is given in Section~\ref{section:qa}.}. Overall, the property of being able to evaluate automatically outweighs these limitations, allowing us to quickly compare different solutions.

\section{Approaching Dialog2API with a Stack of Programs}
\label{section:approach}

We propose a framework for tackling the Dialog2API tasks by representing the dialogue state with a stack of programs. The \textit{program stack} contains the latest version of programs for each goal. The program stack also keeps the status of a program and its execution result. The most recently created/revised program will be on the top of the stack. One example program stack is illustrated in Fig.~\ref{fig:approach}, where we can observe four programs in the stack, one abandoned, two finalized and one drafting. Such a program stack provides useful information for a model to know past interaction, outcomes, current user goal and progress. Also, the order of programs in the stack gives the information on temporal relation among them.

We denote the program stack at turn $t$ with $S_t = \{\{P_1, O_1\}, ..., \{P_N, O_N\}\}$, where $N$ is the stack size. Thus, the program prediction task can be solved by implementing $f(\mathcal{D}, S_{t-1}, U_{<t}, R_{<t}, U_t) \rightarrow P_t$. When program $P_t$ is executed, the execution engine will create $O_t$. After the program stack is updated, the response generation task can be solved with $f(\mathcal{D}, S_t, U_{<t}, R_{<t}, U_t) \rightarrow R_t$.

\subsection{Role of Drafting Programs}
In Dialog2API, we allow a drafting (i.e. pre-execution) program to be generated for a user turn. Note that it is possible for a model to solve the Dialog2API tasks without generating drafting programs (e.g., a model might track the progress of dialogue with latent states and only generate final programs). However, by explicitly presenting the drafting programs in the program stack, the program stack functions as a better summary of dialogue state by telling the goal of a dialogue and current progress towards it. Here is an example of drafting programs that manipulate S3 APIs:
\begin{lstlisting}[language=Python]
# Program #1: drafting
old_path = (*@\textbf{\textcolor{red}{?1}}@*)
new_path = (*@\textbf{\textcolor{red}{?1}}@*).replace(".txt", ".rtf")
s3.copy_object(Bucket="my-bucket", CopySource=f"my-bucket/{new_path}", Key=old_path)
s3.delete_object(Bucket="my-bucket", Key=old_path)
\end{lstlisting}
We can see this drafting program tells that the user's goal is to rename an object by replacing ``.txt'' with ``.rtf'' in the file name. It also indicates that the model already has sufficient information except the object path for finalizing the program. Therefore, with drafting programs, the Dialog2API model can potentially make correct predictions based on only few recent turns, rather than the entire dialogue history.

\paragraph{Placeholders} We allow drafting programs to have placeholders such as ``\textbf{\texttt{?1}}'' in the code, where each placeholder corresponds to a parameter or a block of code. The placeholders enable a drafting program to describe the overall goal and missing information for completing the goal. We require placeholders to be numbered. Placeholders in different locations sharing the same missing information need to share the same number.

\subsection{Program Design Details}
\label{section:program}

\lstset{style=tabstyle}

\begin{table}[t]
\small
\centering
\begin{tabular}{c p{0.85 \linewidth}} \toprule
 & Turn 1 \\
\midrule
$U_1$ & User: Hi, please check the number of files in my bucket  \\ 
$P_1$ & \setbox0=\hbox{\begin{lstlisting}[language=Python]
# Program #1: drafting
objects = s3.list_bucket(Bucket=(*@\textbf{\textcolor{red}{?1}}@*)).get("Contents", [])
return len(objects)
\end{lstlisting}}\mbox{}\box0 \\
$R_1$ & Agent: What is the name of your bucket? \\
\midrule
 & Turn 2 \\
\midrule
$U_2$ & User: The name is zoology-bucket and please check for .txt files  \\ 
$P_2$ & \setbox0=\hbox{\begin{lstlisting}[language=Python]
# Program #1: final
objects = s3.list_bucket(Bucket="zoology-bucket").get("Contents", [])
keys = [obj["Key"] for obj in objects if obj["Key"].endswith(".txt")]
return len(keys)
\end{lstlisting}}\mbox{}\box0 \\
$O_2$ & \setbox0=\hbox{\begin{lstlisting}[language=Python]
10
\end{lstlisting}}\mbox{}\box0 \\
$R_2$ & Agent: You have 10 txt files in ``zoology-bucket'' bucket. \\
\bottomrule
\end{tabular}
\caption{An example dialogue with program annotation for checking the number of certain types of files.}
\label{table:dialog_example}
\end{table}

In this work, we evaluate a program design focusing on providing high functionality and easiness for developers to annotate. In contrast to related work \cite{chen-etal-2020-low,Andreas2020TaskOrientedDA,Yu2019CoSQLAC}, we allow the programs to be generated in standard programming codes. In this work, we ask the model to generate Python codes. We require each program to describe a goal requested by user. It is possible for a user utterance to have multiple goals, thus resulting in multiple programs. The granularity of a goal is controlled by the developer.

\paragraph{Code Structure} Each program is interpreted as a function in the execution engine. A program can return a value with \texttt{return} statement, an empty value \texttt{null} will be returned without such statement. We constrain the return value to be a JSON data. The return value and error generated when executing a program will be observed by the Dialog2API model. In Table~\ref{table:dialog_example}, we show an example dialogue with return statement used in the programs.

Although preferred, we do not require different programs to be annotated with consistent styles. For example, an alternative program can use for loop instead of list comprehension to achieve the same goal described by the program in Table~\ref{table:dialog_example}. 

\paragraph{Goal Revision} In conversational setting, users often add information to or revise a previously mentioned goal during multi-turn interaction. In Dialog2API, when revision happens, we require the model to generate the full updated program status and code\footnote{The whole updated program code needs to be generated, rather than just the revised portion.}. In the example dialogue shown in Table~\ref{table:dialog_example}, the program is revised in Turn 2. The placeholder ``\texttt{?1}'' is replaced by the given bucket name, and an additional line of code is added based on the user request. The program status is updated to ``\texttt{final}'', indicating that the model decided to execute the program. As demonstrated in the program for turn 2, the revision is not limited to the placeholders and it is possible to rewrite or expand the program during revision.

\paragraph{Conversational Exceptions}
\lstset{style=mystyle}
While executing a program $P_t$, if an error is encountered, the error name and message will be recorded in execution result $O_t$. Other than such program-initiated errors, a program can generate conversational exceptions such as the following example:
\begin{lstlisting}[language=Python]
# Program #1: final
raise OutOfScopeRequestError()
\end{lstlisting}
Here, executing this program will generate ``\texttt{OutOfScopeRequestError}'' in the execution result. The error itself does not specify how to generate agent response. The model has to learn the patterns of agent responses from the example dialogues. A full list of conversational exceptions can be found in the appendix \ref{sec:appendix_conversational_exceptions}.

\section{Dataset}

\newcolumntype{C}[1]{>{\arraybackslash}p{#1}}

\begin{table}[t]
\small
\centering
\begin{tabular}{r | r | r} \toprule
\multirow{2}{*}{Dialogues} & \: \: \: \: \: \: \: \: \: number of dialogues & 100 \\
& turns per dialogue & 10.5 \\
\midrule
\multirow{3}{*}{Programs} & average goals in a dialogue & 2.7 \\
& average lines of code & 4.9 \\
& average APIs in a program & 1.4 \\
\midrule
APIs & \multicolumn{2}{C{0.5\linewidth}}{\texttt{GetObject(11\%), PutObject(6\%), ListObjects(32\%), CreateBucket(8\%), CopyObject (10\%), ListBuckets(10\%), DeleteBucket(3\%), DeleteObject(9\%), DeleteObjects(11\%), GetBucketLocation(2\%)}}\\
\bottomrule
\end{tabular}
\caption{Dataset statistics. The last row shows a list of supported API with their ratios in programs.}
\label{table:s3_dataset}
\end{table}

In this work, we collected a dataset that consists of 100 dialogues with turn-level program annotations for manipulating AWS S3 APIs\footnote{\url{https://docs.aws.amazon.com/AmazonS3/latest/userguide/}}. The training/testing split is 46/54. There are no overlapping goals between training and testing dialgoues. For S3 APIs, we can directly reuse the boto3\footnote{\url{https://boto3.readthedocs.io/}} package to build the execution environment. It also satisfies the constraint that the environment change is deterministic through API calls. The environment signature can be easily computed based on a snapshot of s3 buckets and their objects. Details on computing the environment signature is presented in appendix \ref{sec:appendix_signature}. In this dataset, we support 10 APIs that cover around 70\% of the use cases on github\footnote{The coverage is computed based on github search hits of all S3 API usage with boto3.}. The detailed statistics can be found in Table~\ref{table:s3_dataset}. The collection process is done with two phases: (1) dialogue collection and (2) program annotation. We plan to release this dataset.

\paragraph{Dialogue collection} We collect dialogues by first creating the goals. Each goal is described with a canonical utterance (e.g., ``rename all .txt files in zoology-bucket bucket to have .rtf extension''). In the S3 dataset, 50 goals are created and for each goal we collect 2 dialogues. Dialogue collection is done through professional data linguistics.

The dialogues are collected by having two parties simulating the user and the bot with different annotation guidelines. To facilitate the naturalness of dialogues, in each dialogue collection session, the user-side annotator is shown with a canonical utterance as an example goal that they can start with. The user-side annotator then plans the dialogue by considering conversational complexities suggested by the system, however, following the suggested complexities is not a hard requirement. Note that the user-side annotator is blind to the agent-side guideline and supported APIs. As a result, the dialogue may diverge from the canonical utterance either because the dialogue was planned differently, or the user-side annotator observes unexpected agent response.

The agent-side annotators simulate the chatbot by following a stricter guideline. Observing the user utterance, the agent-side annotator is required to first classify whether an conversational exception (e.g., out-of-scope error) should be triggered. For in-domain requests, the agent-side annotator needs to judge whether the program can be completed for the user goal. The annotator should ask for missing information in the order defined by API definitions. When executing final programs, the annotator can predict the outcome of execution, or code the actual program to obtain the results. For each scenario, the guideline gives a guidance for writing the agent response. In average, it takes 6 minutes for collecting a dialogue. In appendix \ref{sec:appendix_example_datapoint}, we show one example dialogue.

\paragraph{Program Annotations} The program annotation is done by first coding and verifying the final programs. Then drafting programs are annotated. As a post-processing step, the final programs will be executed to produce the ground-truth environment signatures as described in Section \ref{sec:eval}.

\begin{table*}[!t]
\small
\centering
\setlength{\tabcolsep}{5pt}
\begin{tabular}{r | c c  c | c c c | c c c} \toprule
 & \multicolumn{3}{c|}{w/ API Doc} & \multicolumn{3}{c|}{w/ In-domain Examples} & \multicolumn{3}{c}{w/ API Doc + Examples} \\
 \midrule
\scriptsize Model & \scriptsize CodeEdits $\uparrow$ & \scriptsize \EMR(\%) & \scriptsize BLEU & \scriptsize CodeEdits$\uparrow$ & \scriptsize \EMR(\%) & \scriptsize BLEU & \scriptsize CodeEdits$\uparrow$ & \scriptsize \EMR(\%) & \scriptsize BLEU \\
\midrule
\codex{ICL w/ Codex Davinci & 23.7 & 8.4 & 15.7 & 14.7 & 30.0  & 28.3 & \textbf{13.2} & \textbf{33.0} & \textbf{29.5}  \\}
ICL w/ Bloom & 40.5 & 0.9 & 5.4 & 38.8 &  7.7 & 14.9 & 27.6 & 4.41 & 10.9  \\
ICL w/ Neox  & 31.2 & 3.6  & 6.8 & 21.3 & 7.5  & 18.2 & 23.5 & 2.59 & 13.1  \\
\bottomrule
\end{tabular}
\caption{Baseline in-context learning with three large language models: \codex{Codex (code-davinci-002),} Bloom (176B) and GPT-Neox (20B).}
\label{table:baseline}
\end{table*}

\section{Experiments}
\subsection{In-context Learning Baseline}
To test the performance of existing models on the proposed Dialog2API tasks, we construct in-context learning baselines with large pre-trained language models (LLMs) \nikos{"following recent line of work [cite]." e.g.  \url{https://arxiv.org/pdf/2203.08568.pdf}}. We follow the approach described in Section~\ref{section:approach} by representing the dialogue state with a stack of programs. For program prediction, we feed the language models the program stack $S_{t-1}$ with the most recent agent and user turns ($R_{t-1}$ and $U_t$ as context. For agent response generation, the prediction is based on $S_t$, $R_{t-1}$ and $U_t$. We also study the impact to performance by including the API definition in context. \nikos{since there is no established way on how to design prompt, is there something we can say about how we came up with the final version? }

\paragraph{Prompt Format} We test three prompt settings:\begin{enumerate}[noitemsep, label=(\alph*)]
    \item API document
    \item In-domain examples
    \item API document with in-domain examples
\end{enumerate}
In setting (a), we prompt the language model with an out-of-domain API document with few input/output examples, followed by API document and test input in the target domain. The out-of-domain examples only provide structural information. 

In setting (b), we show the language models with $K$ examples in the prompt without API document. Each example contains $S_{t-1}, R_{t-1}, U_t, P_t$ or $S_{t}, R_{t-1}, U_t, R_t$ in XML format as follows:
\begin{lstlisting}[language=XML]
<stack>
<goal uid="1" status="drafting">
<program>
s3.copy_object(Bucket=?1, CopySource={"Bucket": ?1, "Key": ?2}, Key=?3)
s3.delete_object(Bucket=?1, Key=?2)
</program>
</goal>
</stack>
<turn>Agent: What is the name of your bucket?</turn>
<turn>User: Can you show me the list of my buckets?</turn>
<output>
<goal uid="2" status="final">
<program>
buckets = []
for bucket in s3.list_buckets().get("Buckets", []):
  buckets.append(bucket["Name"])
return buckets
</program>
</goal>
</output>
\end{lstlisting}\nikos{I'd add captions here to make sure the reader can understand what this is without having to read the whole section. same to other similar inline codes. }
The last example in the prompt is created based on the context, leaving the output section empty for the LLMs to generate.

In setting (c), the in-domain API document is prepended to the examples in setting (b).

\paragraph{Example Retrieval Method} We implement a baseline example retrieval method by computing the similarity scores for dialogue state and history utterances separately. \nikos{let's point to papers that have focused on that } Taking program prediction as an example, denote the example pool as $\{e_1, ..., e_M\}$, where $M$ is the pool size and each example $e_i = \{S_i, R_i, U_i, P_i\}$. The examples in the retrieval pool are extracted from training data. In turn $t$, the similarity score with an example is computed by $\mathrm{score(e_i)} = \mathrm{sim}(S_t, S_i) + \alpha\  \mathrm{sim}(R_{t-1};U_t, R_i;U_i)$. In our experiments, the hyperparameter $\alpha$ is set to $1$.

For computing the similarity score of dialogue states $\mathrm{sim}(S_t, S_i)$, we simply build bag-of-word count vector by counting the keywords including API names and exceptions in the program stack. The cosine similarity of two count vectors is used. For computing the similarity score of utterances $\mathrm{sim}(R_{t-1};U_t, R_i;U_i)$, we concatenate the agent and user utterance. Then we compute the cosine similarity of SBERT \cite{reimers-gurevych-2019-sentence} vectors. Examples of full prompts can be found in appendix \ref{sec:appendix_icl}.
\nikos{Is the SBERT vocabulary preventing us from computing the similarity of dialogue states? if yes we should mention it here so that the choice of simple BOW count is justified. }


\subsection{Baseline Results}

In Table~\ref{table:baseline}, we show the in-context learning baseline results with four language models: GPT Neox, Bloom, Codex Cushman and Codex Davinci. In all experiments, we report results with greedy search. For example-based prompts, we set the number of examples to 5.

We observe that both the Neox and Bloom models can generate programs close to the ground-truth, however, often contain minor mistakes. As \EMR reports the ratio of successful interaction, a mistake in the middle of a dialogue can result in low overall scores. \codex{The Codex model shows considerably strong performance by achieving a \EMR score of 33\% in the best setting.}

Comparing difference prompt settings, we observe that both the API document and in-domain examples benefit the in-context learning baselines. With Codex Davinci, the prompt setting with only API document gives a EMR score of 8.4\%. When the in-domain examples are presented to the language model, adding the API document further increases the EMR by 3\%.


\lstset{style=tabstyle, basicstyle=\ttfamily\scriptsize}
\begin{table}[!t]
\small
\centering
\begin{tabular}{c p{0.85 \linewidth}} \toprule
$U_1$ & User: I need to rename all txt files.  \\ 
$P_1$ & \setbox0=\hbox{\begin{lstlisting}[language=Python]
# Program #1: drafting
objects = s3.list_objects(Bucket=(*@\textbf{\textcolor{red}{?1}}@*)).get("Contents", [])
paths = [obj["Key"] for obj in objects]
paths = [path for path in paths if path.endswith(".txt")]
for path in paths:
  s3.copy_object(
    Bucket=(*@\textbf{\textcolor{red}{?1}}@*),
    CopySource={"Bucket": (*@\textbf{\textcolor{red}{?1}}@*), "Key": path},
    Key=path.replace(".txt", ".txt.bak")
  )
  s3.delete_object(Bucket=(*@\textbf{\textcolor{red}{?1}}@*), Key=path)
\end{lstlisting}}\mbox{}\box0 \\
$R_1$ & Agent: What is the name of your bucket? \\
\midrule
$U_2$ & User: zoology-bucket  \\ 
$P_2$ & \setbox0=\hbox{\begin{lstlisting}[language=Python]
# Program #1: drafting
objects = s3.list_objects(Bucket="zoology-bucket").get("Contents", [])
paths = [obj["Key"] for obj in objects]
paths = [path for path in paths if path.endswith(".txt")]
for path in paths:
  s3.copy_object(
    Bucket="zoology-bucket",
    CopySource={"Bucket": "zoology-bucket", "Key": path},
    Key=path.replace(".txt", ".txt.bak")
  )
  s3.delete_object(Bucket="zoology-bucket", Key=path)
\end{lstlisting}}\mbox{}\box0 \\
$R_2$ & Agent: Can you confirm that you want to delete all .txt files in "zoology-bucket" bucket? \\
\midrule
$U_3$ & User: change the extension from txt to rtf.  \\ 
$P_3$ & \setbox0=\hbox{\begin{lstlisting}[language=Python]
# Program #1: drafting
objects = s3.list_objects(Bucket="zoology-bucket").get("Contents", [])
paths = [obj["Key"] for obj in objects]
paths = [path for path in paths if path.endswith(".txt")]
for path in paths:
  s3.copy_object(
    Bucket="zoology-bucket",
    CopySource={"Bucket": "zoology-bucket", "Key": path},
    Key=path.replace(".txt", ".rtf")
  )
  s3.delete_object(Bucket="zoology-bucket", Key=path)
\end{lstlisting}}\mbox{}\box0 \\
$R_3$ & Agent: Can you confirm that you want to change the extension from txt to rtf in "zoology-bucket" bucket? \\
\midrule
$U_4$ & User: Yes  \\ 
$P_4$ & \setbox0=\hbox{\begin{lstlisting}[language=Python]
# Program #1: final
...
\end{lstlisting}}\mbox{}\box0 \\
$O_4$ & \setbox0=\hbox{\begin{lstlisting}[language=Python]
null
\end{lstlisting}}\mbox{}\box0 \\
$R_4$ & Agent: Your query is successfully executed. \\
\bottomrule
\end{tabular}
\caption{An example dialogue generated by in-context learning baseline\codex{ w/ Codex model}. The actual program code for $P_4$ is omitted here as there is no revision in the code from $P_3$.}
\label{table:qa_example}
\end{table}

\subsection{Qualitative Analysis}
\label{section:qa}

In Table~\ref{table:qa_example}, we show a test dialogue generated by the in-context learning baseline\codex{ with Codex Davinci} \cite{Chen2021EvaluatingLL}. The example shown in the table is overall a successful dialogue by executing a correct program in the last turn. However, it contains several typical error patterns. The program generated in the first turn is almost correct, however, it attempts to rename txt files to have ``\texttt{.txt.bak}'' extension. This is a typical hallucination pattern caused by the language model. In turn 3, as the user explicitly requests to change the extension to ``\texttt{.rtf}'', the hallucination is fixed during program revision.

In turn 2, the model attempts to generate a confirmation message. However, the generated response failed to correctly describe the purpose of the program. When generating test dialogues for computing \EMR, as we always feed the model with the ground-truth user utterance in corresponding turns, the user utterance may not fit into the context when the generated agent response contains error. This can result in over-estimated \EMR score.

In the last turn, as a correct program is executed, the environment signature matches the ground-truth, resulting an \EMR score of 1.0 for this test dialogue. However, the mistake in agent response $R_2$ leads to a lowered response BLEU score.

\section{Related Work}
The closest work to this paper is \citet{Andreas2020TaskOrientedDA}, where the task-oriented dialogue is formulated as data flow synthesis problem. There are two major differences comparing to our work. First, we focus on evaluating the model capability of generating a functional chatbot based on given API documents and few example dialogues. Second, in our proposed approach, we represent the dialogue state with a program stack instead of the dataflow in aforementioned work. We allow much more flexible goal description though free-form programming codes. When the user goal is revised, unlike the approach that generates a sequence of actions for revision, we require the full updated program to be generated. As a consequence, our approach is difficult to be evaluated with exact match, which is mitigated with an execution-based metric.

Comparing to semantic parsing works \cite{Yu2019CoSQLAC,Gupta2018SemanticPF}, the model in Dialog2API needs to deal with structurally complex goals. In Dialog2API, as we are focusing on few-shot setting, the model is required to generalize and learn to generate programs with unseen structure and combination of APIs. Additionally, our work is related to program synthesis with large language models \cite{Austin2021ProgramSW,Wang2021CodeT5IU}. \citet{Cheng2022BindingLM} incorporates Codex for parsing natural language request into executable queries, where a neural-based function can be inserted for solving commonsense QA.  \citet{Nijkamp2022CodeGenAO} further trained a large model to generate programs in a multi-turn fashion. In contrast to these work, we require the model to analyze the underlying user goals to generate corresponding programs based on dialogue context. We further require the model to generate responses for interaction.

\section{Conclusion}
In this work, we introduce Dialog2API for greatly expanding the functionality of TOD systems by learning to generate programs triggering a set of APIs. Such a paradigm has the potential to solve complex tasks by interacting with human users. To build the chatbot, we ask the developers to provide API documents, and optionally example dialogues with program annotation to customize the dialogue experience.

To evaluate the program prediction performance, we propose to measure the execution match ratio (\EMR) by matching the environment signature in an end-to-end fashion. This metric reflects the user experience when conversing with the system.

We propose an approach for tackling the Dialog2API by representing the dialogue state with a stack of programs, and condition the program/response generation with API documents, the program stack, and dialogue history. Following this approach, the baseline results on our constructed dataset for manipulating S3 APIs show that the best existing large language model can achieve a \EMR score of $33\%$, with $22\%$ of test dialogue being perfectly generated (i.e., \EMR=$1.0$).  We also observe that both the API documents and example dialogues are helpful for predicting more correct programs.

Dialog2API has the potential to provide seamless dialogue experience. The dialogue is not restricted by the underlying schema (e.g., commands, queries, intents or APIs). The core model is required to generate programs that combines different APIs in a way that fits in to the dialogue context. By bootstrapping from the API document and examples, we allow the dialogue system to be created with minimal efforts.  


\bibliography{anthology,custom}

\newpage
\appendix

\section{Comparing dialogue experience of different paradigms}
\label{sec:appendix_paradigm_examples}

In following snippets, we show example dialogues provided by different paradigms to highlight the difference:
\lstset{style=mystyle}

- (a) Utterance-based semantic parsing
\begin{lstlisting}
User: Create a bucket named "sea-creatures" in "us-west-2" region.
Bot: The bucket "sea-creatures" is created.
User: List all files in "zoology-bucket" bucket.
Bot: Here are the files:
 - land_animals/pinniped.txt
 - land_animals/primate.txt
 - sea_animals/dolphin.txt
 - sea_animals/otter.txt
User: Copy "sea_animals/dolphin.txt" from "zoology-bucket" bucket to "sea-creatures" bucket.
Bot: Done.
User: Copy "sea_animals/otter.txt" from "zoology-bucket" bucket to "sea-creatures" bucket.
Bot: Done.
\end{lstlisting}

- (b) Slot-filling
\begin{lstlisting}
User: Create a bucket
Bot: What is the bucket name?
User: sea-creatures.
Bot: In which region do you want to create the bucket?
User: us-west-2
Bot: Your bucket "sea-creatures" is created.
User: List files in "zoology-bucket" bucket.
Bot: Here are the files:
 - land_animals/pinniped.txt
 - land_animals/primate.txt
 - sea_animals/dolphin.txt
 - sea_animals/otter.txt
User: Copy files from "zoology-bucket" bucket to "sea-creatures" bucket.
Bot: Which file do you want to copy?
User: sea_animals/dolphin.txt.
Bot: File is copied. Do you want to copy another file?
User: Yes
Bot: Which file do you want to copy?
User: sea_animals/otter.txt.
Bot: File is copied. Do you want to copy another file?
User: No
Bot: Please let me know if there is anything I can help.
\end{lstlisting}

- (c) Dialog2API
\begin{lstlisting}
User: Create a bucket, call it "sea-creatures".
Bot: In which region do you want to create the bucket?
User: us-west-2
Bot: Your bucket "sea-creatures" is created.
User: I want to copy some files from  "zoology-bucket" bucket to the bucket I just created.
Bot: Which files do you want to copy?
User: All files with "sea" in filename.
Bot: Can you confirm that you want to copy all files in "zoology-bucket" bucket containing "sea" in filename to "sea-creatures" bucket?
User: Yes.
Bot: Your query is executed. Following files are copied:
 - sea_animals/dolphin.txt
 - sea_animals/otter.txt
\end{lstlisting}

Here, the chatbot supports three different functions in the schema: CreateBucket, ListObjects and CopyObject. The dialogues are constructed by assuming the user has the knowledge on how to use the chatbot and the chatbot perfectly understands the user request.

We can see that the utterance-based semantic parsing approach typically requires the user request to follow the underlying schema and functions closely, so that the request can be correctly parsed. However, there is a steep learning curve for a user to be able to use such kind of system. The slot-filling system surfers from the same problem. The dialogue flow is restricted by the intents and slots defined in the system. In Dialog2API paradigm, as the model can write the following program that combines different APIs, it allows the user to interact with the system in a natural way:
\begin{lstlisting}[language=Python]
# Program #2: drafting
keys = ListObjects(BucketName="zoology-bucket")
copied_objects = []
for key in keys:
    if "sea" in key:
        CopyObject(
            CopySource="zoology-bucket/{}".format(key),
            Bucket="sea-creatures",
            Key=key
        )
        copied_objects.append(key)
return copied_objects
\end{lstlisting}

\section{Computing environment signature}
\label{sec:appendix_signature}

The environment signature is a checksum that summarizes the environment state and program return value. In s3 dataset, the checksum is computed based on following information:
\begin{itemize}
    \item Bucket names in the account
    \item All object keys in each bucket
    \item Content of each object
    \item Program return value after execution (if applicable)
\end{itemize}

\section{Conversational Exceptions}
\label{sec:appendix_conversational_exceptions}

In the program design for the S3 dataset, we allow the following conversational exceptions:
\begin{itemize}
    \item \textbf{OutOfScopeRequestError}: the user request is out of the scope of the target domain
    \begin{itemize}
        \item Example: Can I switch to a different AWS account?
    \end{itemize}
    \item \textbf{AmbiguousRequestError}: the user request is considered as in-domain, how ever it's ambiguous
    \begin{itemize}
        \item Example: I want to do something with a bucket.
    \end{itemize}
    \item \textbf{ChitChat}: the user request is chitchat
    \begin{itemize}
        \item Example: Good, nice job
    \end{itemize}
    \item \textbf{FAQ}: the user request is an in-domain FAQ question, however couldn't be solved by a program
    \begin{itemize}
        \item Example: Can I put one bucket into another bucket?
    \end{itemize}
    \item \textbf{OverSpecificationError}: the user request contains more constraints that are impossible to generate suitable programs
    \begin{itemize}
        \item Example: I want to create a new bucket and mark it with red color.
    \end{itemize}
    \item \textbf{EndDialog}: the user request is to end the dialogue
    \begin{itemize}
        \item Example: Thank you so much today
    \end{itemize}
\end{itemize}

\section{Example Dialogue with Program Annotation}
\label{sec:appendix_example_datapoint}

In this section, we show an example dialogue in the testing set with turn-level program annotation:

\begin{lstlisting}[language=XML, basicstyle=\ttfamily\scriptsize]
<conversation uid="ListObjects_10">
<metadata>
<initial_signature>dac8c078</initial_signature>
</metadata>
<turns>
<turn>Agent: How can I help you?</turn>
<turn>User: can you list all the txt files inside the 'mammals' subfolder which is inside the 'land_animals' folder</turn>
<goal uid="1" status="drafting">
<program>
    objects = s3.list_objects(Bucket=?1, Prefix="land_animals/mammals").get("Contents", [])
    paths = [obj["Key"] for obj in objects]
    paths = [path for path in paths if path.endswith(".txt")]
    return paths
  </program>
</goal>
<turn>Agent: What is the name of your bucket?</turn>
<turn>User: Can you check it for me? The standard bucket I use that has 'land_animals' folder</turn>
<goal uid="2" status="final">
<program>
    raise OutOfScopeRequestError()
  </program>
<result>null</result>
<error>{"error": "OutOfScopeRequestError", "message": ""}</error>
<signature>da12ac22</signature>
</goal>
<turn>Agent: Sorry, I cannot search for a bucket.</turn>
<turn>User: Can you then list the buckets I have</turn>
<goal uid="3" status="final">
<program>
    buckets = s3.list_buckets().get("Buckets", [])
    return [bucket["Name"] for bucket in buckets]
  </program>
<result>["zoology-bucket"]</result>
<error/>
<signature>cb7a0702</signature>
</goal>
<turn>
Agent: Here are your buckets:
- zoology-bucket
</turn>
<turn>User: zoology-bucket then</turn>
<goal uid="1" status="final">
<program>
    objects = s3.list_objects(Bucket="zoology-bucket", Prefix="land_animals/mammals").get("Contents", [])
    paths = [obj["Key"] for obj in objects]
    paths = [path for path in paths if path.endswith(".txt")]
    return paths
  </program>
<result>["land_animals/mammals/bat.txt", "land_animals/mammals/deer.txt", "land_animals/mammals/pika.txt"]</result>
<error/>
<signature>7139c614</signature>
</goal>
<turn>
Agent: Here are your objects:
- land_animals/mammals/bat.txt
- land_animals/mammals/deer.txt
- land_animals/mammals/pika.txt
</turn>
<turn>User: cool! Can you display these files in the reverse space order (largest files first)</turn>
<goal uid="5" status="final">
<program>
    objects = s3.list_objects(Bucket="zoology-bucket", Prefix="land_animals/mammals").get("Contents", [])
    pairs = [(obj["Key"], obj["Size"]) for obj in objects if obj["Key"].endswith(".txt")]
    pairs.sort(key=lambda pair: pair[1], reverse=True)
    return pairs
  </program>
<result>[["land_animals/mammals/bat.txt", 1551], ["land_animals/mammals/pika.txt", 878], ["land_animals/mammals/deer.txt", 402]]</result>
<error/>
<signature>79b9d303</signature>
</goal>
<turn>
Agent: Here are your objects:
- land_animals/mammals/bat.txt 1551 bytes
- land_animals/mammals/pika.txt 878 bytes
- land_animals/mammals/deer.txt 402 bytes
</turn>
<turn>User: all good, thank you</turn>
<goal uid="6" status="final">
<program>
    raise EndDialog()
  </program>
<result>null</result>
<error>{"error": "EndDialog", "message": ""}</error>
<signature>d0eeb78a</signature>
</goal>
<turn>Agent: Thanks, please let me know if there is anything I can do for you.</turn>
</turns>
</conversation>
\end{lstlisting}

\section{In-context Learning Details}
\label{sec:appendix_icl}

In this section, we report more technical details for in-context learning experiments. For the program prediction and response generation tasks, we prompt the large language model separately, however with same prompting format and retrieval strategy.

\subsection{Prompt Example for Program Prediction}
Here we show an example prompt we send to the language model for program prediction task in the ``API Doc + Example'' setting:

\begin{lstlisting}[language=XML, basicstyle=\ttfamily\scriptsize]
<def>
def get_object(self, Bucket: str, Key: str):
"""Retrieves objects from Amazon S3."""

def put_object(self, Bucket: str, Key: str, Body: Optional[str] = ""):
"""Adds an object to a bucket."""

def delete_object(self, Bucket: str, Key: str):
"""Delete an object from a bucket."""

def list_objects(self, Bucket: str, Prefix: Optional[str] = "", Delimiter: Optional[str] = ""):
"""Returns some or all of the objects in a bucket."""

def create_bucket(self, Bucket: str, CreateBucketConfiguration: Dict):
"""Creates a new S3 bucket."""

def copy_object(self, Bucket: str, CopySource: Union[Dict, str], Key: str):
"""Creates a copy of an object that is already stored in Amazon S3."""

def list_buckets(self):
"""Returns a list of all buckets owned by the authenticated sender of the request."""

def delete_bucket(self, Bucket: str):
"""Deletes the S3 bucket."""

def delete_objects(self, Bucket: str, Delete: Dict):
"""Delete multiple objects from a bucket."""

def get_bucket_location(self, Bucket: str):
"""Returns the Region the bucket resides in."""

def head_bucket(self, Bucket: str):
"""Determine if a bucket exists and you have permission to access it."""
</def>

<stack>
<goal uid="1" status="drafting">
<program>
    s3.copy_object(
      Bucket=?1,
      CopySource={"Bucket": ?1, "Key": ?2},
      Key=?3
    )
    s3.delete_object(
      Bucket=?1,
      Key=?2
    )
  </program>
</goal>
</stack>
<turn>Agent: What is the name of your bucket?</turn>
<turn>User: Can you show me the list of my buckets?</turn>
<output>
<goal uid="2" status="final">
<program>
    buckets = []
    for bucket in s3.list_buckets().get("Buckets", []):
      buckets.append(bucket["Name"])
    return buckets
  </program>
</goal>
</output>

<stack>
<goal uid="1" status="drafting">
<program>
    s3.delete_objects(
      Bucket=?1
      Delete={"Objects": ?2}
    )
  </program>
</goal>
</stack>
<turn>Agent: What is the name of your bucket?</turn>
<turn>User: zoology-bucket</turn>
<output>
<goal uid="1" status="drafting">
<program>
    s3.delete_objects(
      Bucket="zoology-bucket"
      Delete={"Objects": ?2}
    )
  </program>
</goal>
</output>

<stack>
<goal uid="1" status="drafting">
<program>
    paths = [obj["Key"] for obj in s3.list_objects(Bucket=?1, Prefix="sea_animals/").get("Contents", [])]
    if paths:
      s3.delete_objects(
        Bucket=?1,
        Delete={"Objects": [{"Key": path} for path in paths]}
      )
  </program>
</goal>
</stack>
<turn>Agent: What is the name of your bucket?</turn>
<turn>User: zoology-bucket</turn>
<output>
<goal uid="1" status="drafting">
<program>
    paths = [obj["Key"] for obj in s3.list_objects(Bucket="zoology-bucket", Prefix="sea_animals/").get("Contents", [])]
    if paths:
      s3.delete_objects(
        Bucket="zoology-bucket",
        Delete={"Objects": [{"Key": path} for path in paths]}
      )
  </program>
</goal>
</output>

<stack>
<goal uid="1" status="drafting">
<program>
    objects = s3.list_objects(
        Bucket=?1
    ).get("Contents", [])
    objects_to_delete = [{"Key": obj["Key"]} for obj in objects]
    if objects_to_delete:
        s3.delete_objects(
            Bucket=?1,
            Delete={
                "Objects": objects_to_delete
            }
        )
    s3.delete_bucket(
        Bucket=?1
    )
  </program>
</goal>
</stack>
<turn>Agent: What is the name of the bucket?</turn>
<turn>User: "zoology-bucket"</turn>
<output>
<goal uid="1" status="drafting">
<program>
    objects = s3.list_objects(
        Bucket="zoology-bucket"
    ).get("Contents", [])
    objects_to_delete = [{"Key": obj["Key"]} for obj in objects]
    if objects_to_delete:
        s3.delete_objects(
            Bucket="zoology-bucket",
            Delete={
                "Objects": objects_to_delete
            }
        )
    s3.delete_bucket(
        Bucket="zoology-bucket"
    )
  </program>
</goal>
</output>

<stack>
<goal uid="1" status="drafting">
<program>
    objects = s3.list_objects(
        Bucket="zoology-bucket"
    ).get("Contents", [])
    objects_to_delete = [{"Key": obj["Key"]} for obj in objects]
    if objects_to_delete:
        s3.delete_objects(
            Bucket="zoology-bucket",
            Delete={
                "Objects": objects_to_delete
            }
        )
    s3.delete_bucket(
        Bucket="zoology-bucket"
    )
  </program>
</goal>
</stack>
<turn>Agent: Can you confirm that you want to delete "zoology-bucket" bucket?</turn>
<turn>User: Yes.</turn>
<output>
<goal uid="1" status="final"/>
</output>

<stack>
<goal uid="1" status="drafting">
<program>
    s3.copy_object(
      Bucket=?1,
      CopySource={"Bucket": ?1, "Key": ?2},
      Key=?3
    )
    s3.delete_object(
      Bucket=?1,
      Key=?2
    )
  </program>
</goal>
</stack>
<turn>Agent: What is the name of your bucket?</turn>
<turn>User: it's zoology-bucket</turn>
<output>
\end{lstlisting}

\subsection{Prompt Example for Response Generation}
Here we show an example prompt we send to the language model for response generation task:
\begin{lstlisting}[language=XML, basicstyle=\ttfamily\scriptsize]
<stack>
<goal uid="1" status="final">
<program>
    objects = s3.list_objects(Bucket="zoology-bucket", Prefix="mammals/").get("Contents", [])
    paths = [obj["Key"] for obj in objects]
    return paths
  </program>
<result>[]</result>
<error/>
</goal>
</stack>
<turn>Agent: What is the name of your bucket?</turn>
<turn>User: zoology-bucket</turn>
<output>
<turn>Agent: You do not have any object with prefix "mammals/" in "zoology-bucket" bucket.</turn>
</output>

<stack>
<goal uid="1" status="final">
<program>
    objects = s3.list_objects(Bucket="zoology bucket", Prefix="sea_animals/").get("Contents", [])
    paths = [obj["Key"] for obj in objects]
    return paths
  </program>
<result>null</result>
<error>{"error": "NoSuchBucket", "message": "An error occurred (NoSuchBucket) when calling the ListObjects operation: Your previous request to create the named bucket succeeded and you already own it."}</error>
</goal>
</stack>
<turn>Agent: What is the name of your bucket?</turn>
<turn>User: zoology bucket</turn>
<output>
<turn>Agent: Sorry, the bucket name "zoology bucket" is invalid.</turn>
</output>

<stack>
<goal uid="1" status="final">
<program>
    objects = s3.list_objects(Bucket="zoology-bucket", Prefix="land_animals/").get("Contents", [])
    paths = [obj["Key"] for obj in objects]
    return paths
  </program>
<result>["land_animals/mammals/bat.txt", "land_animals/mammals/deer.txt", "land_animals/mammals/pika.txt", "land_animals/pinniped.txt", "land_animals/primate.txt"]</result>
<error/>
</goal>
</stack>
<turn>Agent: What is the name of your bucket?</turn>
<turn>User: "zoology-bucket"</turn>
<output>
<turn>Agent: Here are your objects:
- land_animals/mammals/bat.txt
- land_animals/mammals/deer.txt
- land_animals/mammals/pika.txt
- land_animals/pinniped.txt
- land_animals/primate.txt</turn>
</output>

<stack>
<goal uid="1" status="final">
<program>
    objects = s3.list_objects(Bucket="zoology-bucket").get("Contents", [])
    paths = [obj["Key"] for obj in objects]
    return "land_animals/insects.txt" in paths
  </program>
<result>false</result>
<error/>
</goal>
</stack>
<turn>Agent: What is the name of your bucket?</turn>
<turn>User: in the "zoology-bucket".</turn>
<output>
<turn>Agent: No, I cannot find this object.</turn>
</output>

<stack>
<goal uid="1" status="final">
<program>
    objects = s3.list_objects(Bucket="zoology-bucket").get("Contents", [])[:3]
    paths = [obj["Key"] for obj in objects]
    return paths
  </program>
<result>["land_animals/mammals/bat.txt", "land_animals/mammals/deer.txt", "land_animals/mammals/pika.txt"]</result>
<error/>
</goal>
</stack>
<turn>Agent: What is the name of your bucket?</turn>
<turn>User: zoology-bucket. i am curious what are the contents of this bucket</turn>
<output>
<turn>
Agent: Here are the objects:
- land_animals/mammals/bat.txt
- land_animals/mammals/deer.txt
- land_animals/mammals/pika.txt
</turn>
</output>

<stack>
<goal uid="1" status="final">
<program>
    objects = s3.list_objects(Bucket="zoology-bucket").get("Contents", [])
    paths = [obj["Key"] for obj in objects]
    paths = [path for path in paths if path.endswith(".txt")]
    for path in paths:
      s3.copy_object(
        Bucket="zoology-bucket",
        CopySource={"Bucket": "zoology-bucket", "Key": path},
        Key=path.replace(".txt", ".txt.bak")
      )
  </program>
<result>null</result>
<error/>
</goal>
</stack>
<turn>Agent: What is the name of your bucket?</turn>
<turn>User: zoology-bucket</turn>
<output>
\end{lstlisting}

\end{document}